\documentclass{article}

\usepackage[nonatbib,preprint]{neurips_2024}

\usepackage[utf8]{inputenc} 
\usepackage[T1]{fontenc}    
\usepackage{hyperref}       
\usepackage{url}            
\usepackage{booktabs}       
\usepackage{amsfonts}       
\usepackage{nicefrac}       
\usepackage{microtype}      
\usepackage{xcolor}         
\usepackage[pdftex]{graphicx}
\usepackage{algorithm} 
\usepackage{algpseudocode} 
\usepackage{wrapfig}
\usepackage{enumitem}
\usepackage{authblk}

\author[*, +]{Vinay Joshi}
\author[*]{Prashant Laddha}
\author[*]{Shambhavi Sinha}
\author[*]{Om Ji Omer}
\author[*]{Sreenivas Subramoney}
\affil[*]{Intel Labs}
\affil[+]{corresponding author: vinay.joshi@intel.com}

\newcommand{\bt}[1]{\textcolor{black}{#1}}

\begin{document}

\title{QCQA: \underline{Q}uality and \underline{C}apacity-aware grouped \underline{Q}uery \underline{A}ttention}
\maketitle

\begin{abstract}
Excessive memory requirements of key and value features (KV-cache) present significant challenges in the autoregressive inference of large language models (LLMs), restricting both the speed and length of text generation.
Approaches such as Multi-Query Attention (MQA) and Grouped Query Attention (GQA) mitigate these challenges by grouping query heads and consequently reducing the number of corresponding key and value heads. 
However, MQA and GQA decrease the KV-cache size requirements at the expense of LLM accuracy (quality of text generation). 
These methods do not ensure an optimal tradeoff between KV-cache size and text generation quality due to the absence of quality-aware grouping of query heads. 
To address this issue, we propose Quality and Capacity-Aware Grouped Query Attention (QCQA), which identifies optimal query head groupings using an evolutionary algorithm with a computationally efficient and inexpensive fitness function.
We demonstrate that QCQA achieves a significantly better tradeoff between KV-cache capacity and LLM accuracy compared to GQA.
For the Llama2 $7\,$B model, QCQA achieves $\mathbf{20}$\% higher accuracy than GQA with similar KV-cache size requirements in the absence of fine-tuning.
After fine-tuning both QCQA and GQA, for a similar KV-cache size, QCQA provides $\mathbf{10.55}\,$\% higher accuracy than GQA. 
Furthermore, QCQA requires $40\,$\% less KV-cache size than GQA to attain similar accuracy.
The proposed quality and capacity-aware grouping of query heads can serve as a new paradigm for KV-cache optimization in autoregressive LLM inference.

\end{abstract}

\section{Introduction}
\label{sec:intro}
Large language models (LLMs) have achieved remarkable performance across various natural language processing tasks \cite{gpt-3, openai2024gpt4, touvron2023llama, jiang2023mistral, anil2023palm, radford2019language, zhang2022opt}. Consequently, they are widely used in real-time applications such as programming \cite{jiang2024mixtral}, chatbots \cite{gpt-3}, and content creation, including image, text, and music generation \cite{esser2024scaling, bellagente2024stable, yuan2024chatmusician}. Autoregressive inference, which underpins these applications, requires caching past key and value features (KV-cache) for future token computation \cite{gpt-3}. 
The KV-cache size increases linearly with the mini-batch size, input sequence length, number of heads per layer, and number of layers. 
The growing complexity of language modeling tasks demands LLMs with a higher number of attention layers and heads per layer to process longer input sequence lengths. 
This leads to a \bt{significant increase in} KV-cache, often exceeding the size of the model parameters, especially when the mini-batch size is increased to improve throughput. 
\bt{Memory accesses to large KV-cache becomes the primary bottleneck in autoregressive LLM inference \cite{zhang2024kv}, adversely impacting the latency, throughput, and energy consumed for token generation.} 
Therefore, optimizing KV-cache capacity is essential for achieving efficient, low-latency, and energy-efficient autoregressive LLM inference.

\begin{wrapfigure}{r}{0.5\textwidth}
    \centering
    \includegraphics[width=0.45\textwidth]{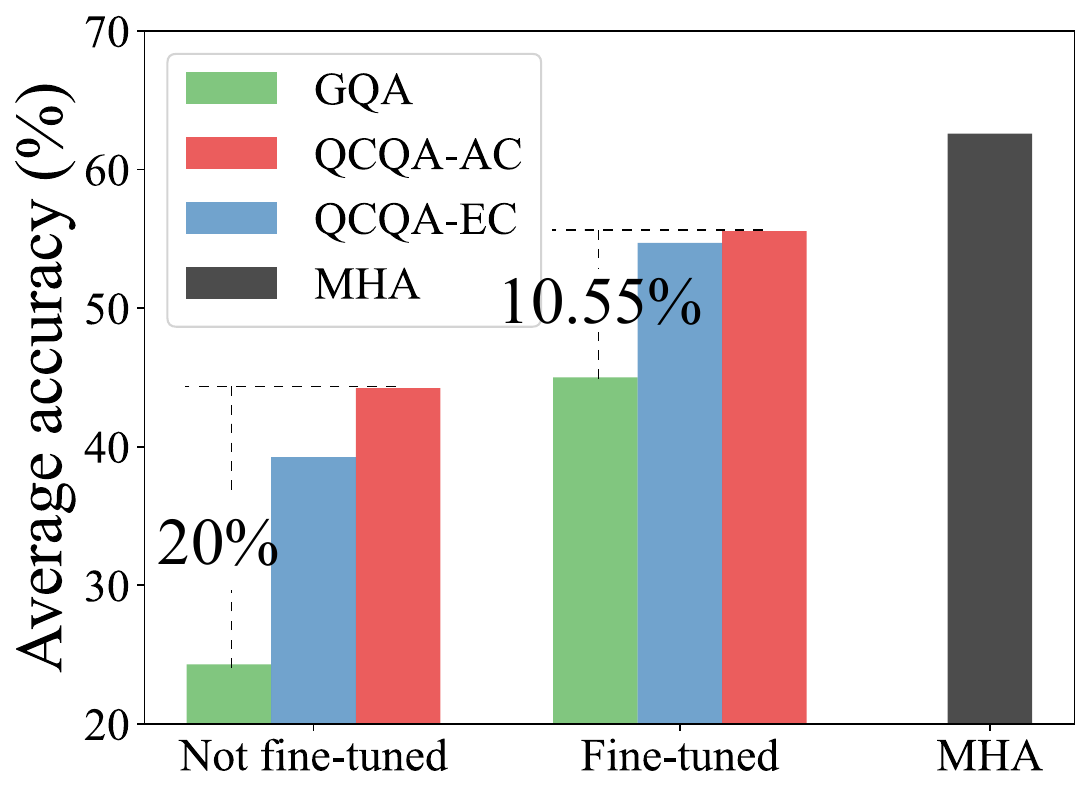}
    \caption{The average accuracy of GQA and QCQA at $50\,$\% of the original KV-cache size. For QCQA we allow group cardinality to be either equal (QCQA-EC) or arbitrary (QCQA-AC). QCQA-AC outperforms QCQA-EC and GQA, and \textbf{without fine-tuning QCQA-AC performs the same as that of GQA fine-tuned for 3 epochs}. }
    \label{fig:into_fig}    
\end{wrapfigure}
Several techniques have been proposed to optimize KV-cache size requirements, including reduced precision \cite{hooper2024kvquant}, retaining only significant keys and values \cite{delcorro2023skipdecode, zhang2023h2o, anagnostidis2023dynamic, liu2023scissorhands}, and grouped attention, which involves sharing key and value features across multiple query heads \cite{ainslie2023gqa, shazeer2019fast_mqa}. This paper focuses on grouped attention for KV-cache optimization.
Two prominent grouped attention methods are 1) Multi-Query Attention (MQA) and 2) Grouped-Query Attention (GQA). 
MQA \cite{shazeer2019fast_mqa} consolidates all query heads into a single group, using one key (and value) head per layer. 
GQA \cite{ainslie2023gqa}, a generalization of MQA, divides query heads into multiple groups of equal size, with each group sharing a single key (and value) head. 
GQA has witnessed widespread adoption in popular LLM models such as Llama 2 \cite{touvron2023llama}, Llama 3 \cite{llama3}, Mistral \cite{jiang2023mistral}, Gemma \cite{gemmateam2024gemma}, Starcoder \cite{lozhkov2024starcoder}, and Meditron \cite{MEDITRON-70B}. 
MQA proposed full training \bt{while} GQA requires additional retraining (uptraining) for $5-10\,$\% of training steps \bt{compared to full training}\cite{ainslie2023gqa} to reduce the loss in accuracy. 
The time and resource-intensive training required in MQA and GQA can also be prohibitive in many cases.

MQA and GQA cannot guarantee an optimal tradeoff between KV-cache size and LLM accuracy because of- \textbf{1)} lack of quality and capacity-aware grouping of query heads, \textbf{2)} the constraint of equally sized groups of consecutive heads could force two heads with highly distinct distribution into the same group. 
Determining the optimal grouping through brute force evaluation of all possible combinations \bt{of grouping} is \bt{impractical} due to the enormous number of \bt{candidate groupings}, represented by the Stirling number of the second kind $S(H, P)$ \cite{stirling-numbers} for $H$ number of heads and $P$ groups. For instance, in the Llama2 7B model, arranging 32 heads into 4 groups can be done in $S(32, 4) \approx 7 \times 10^{17}$ ways. Beyond the vast search space, the computational expense of evaluating LLM accuracy \bt{for each candidate grouping} presents a significant challenge \cite{polo2024tinybenchmarks}.


\bt{We propose Quality and Capacity-Aware Grouped Query Attention (QCQA) which minimizes both accuracy loss and KV-cache capacity through a careful choice of layers and query heads for grouping.}
Unlike MQA and GQA, the QCQA can create arbitrarily sized groups of query heads using an evolutionary algorithm. To circumvent the need for expensive LLM accuracy computations, QCQA employs a simple and computationally efficient fitness function that reliably indicates potential accuracy loss. 
\bt{A comparison of average accuracy of different grouping techniques is shown in Figure~\ref{fig:into_fig}.}
 Our results show that, for the Llama2 $7\,$B model, 
 QCQA attains {$\mathbf{20}$\%} higher accuracy than GQA with similar KV-cache size requirements in the absence of fine-tuning.
After fine-tuning both QCQA and GQA, for similar KV-cache size, QCQA provides $\mathbf{10.55}\,$\% more accuracy than GQA. 
\bt{Furthermore, QCQA requires $\mathbf{40}\,$\% lesser KV-cache size compared to GQA for the same accuracy.}

\textbf{Our major contributions} are:
\begin{enumerate}[nolistsep]
    \item We propose a quality and capacity-aware grouping of query heads within and across layers for an optimal tradeoff between LLM accuracy and KV-cache size.
    \item We formulate two unique representations for applying an evolutionary algorithm to form groups of query heads with arbitrary or equal cardinality.
    \item We eliminate the need for expensive LLM evaluations by proposing a simple and inexpensive fitness estimate, weight-sharing error, as a reliable indicator of LLM accuracy drop.
\end{enumerate}

\section{Background}

\subsection{LLM inference acceleration}

Autoregressive inference in LLMs is often impacted by the repetitive loading of key and value features for computing the next output token.
Uncontrolled growth in KV-cache due to increasing mini-batch size or input sequence length could exceed the size of LLM parameters (weights) per se.
Researchers have explored several approaches to optimize KV-cache memory requirements such as using low precision for key and value features \cite{hooper2024kvquant}, KV-cache eviction strategies \cite{liu2023scissorhands} to retain only important tokens, token level early exit strategy \cite{delcorro2023skipdecode}, and sharing-based attention acceleration \cite{ainslie2023gqa, shazeer2019fast_mqa}. 
In this work, we explore sharing-based attention acceleration using an evolutionary algorithm to find an optimal tradeoff between KV-cache memory requirements and LLM evaluation accuracy.

\textbf{Sharing-based attention acceleration} shrinks head dimension by arranging multiple Query heads in different groups and often mean-pools corresponding key and value heads. 
For a mini-batch size $B$, input sequence length $T$, and $H$ number of heads both key and value features per attention layer have dimension key~$\in~\mathbb{R}^{[B,T,H,d_k]}$ and value~$\in~\mathbb{R}^{[B,T,H,d_v]}$, respectively. 
The head dimension of model weights corresponding to key ($W_{K} \in \mathbb{R}^{[H,d_k,d_{model}]}$) and value ($W_{V} \in \mathbb{R}^{[H,d_v,d_{model}]}$) features is reduced to shrink the KV-cache size.
$d_k$, $d_v$, and $d_{model}$ are feature dimensions of key, value, and model, respectively. 
Either of the four dimensions can be shrunk to reduce the overall KV-cache memory requirements. 
However, reducing the mini-batch size impacts the throughput of LLM execution on the hardware. 
Decreasing the number of input tokens affects the model's performance on longer sequences. 
Hence a more natural choice is to reduce the KV-cache capacity by decreasing the number of heads.

\subsection{Evolutionary algorithms (EA) for architecture search}
Shrinking of head dimension for LLM inference acceleration bears some similarities to neural architecture search (NAS) \cite{enas} commonly used in optimizing deep neural network model architecture.
The problem of arranging Query heads in distinct groups has an exhaustively large search space (see section \ref{sec:intro}). 
In \cite{cummings2022hardwareaware, Wang_2020, Lu_9328602, laube2022what} authors have shown promising results in using EA-based NAS for convolution neural network (CNN) architecture search to enable efficient deployment as proposed in \cite{ofa}. 
A major bottleneck in using EA for architecture search is expensive candidate evaluations due to training or inference.
Previous research \cite{cummings2022hardwareaware, Wang_2020, Lu_9328602, laube2022what} has employed predictor functions to estimate the accuracy of a given CNN architecture to eliminate expensive inference runs, albeit at the cost of running expensive inference to collect data for supervised training of a predictor function.

\section{\textbf{Q}uality and \textbf{C}apacity Aware grouped \textbf{Q}uery \textbf{A}ttention (QCQA)}
\label{sec:methodology}
\begin{figure}
    \centering
    \includegraphics[width=0.95\linewidth]{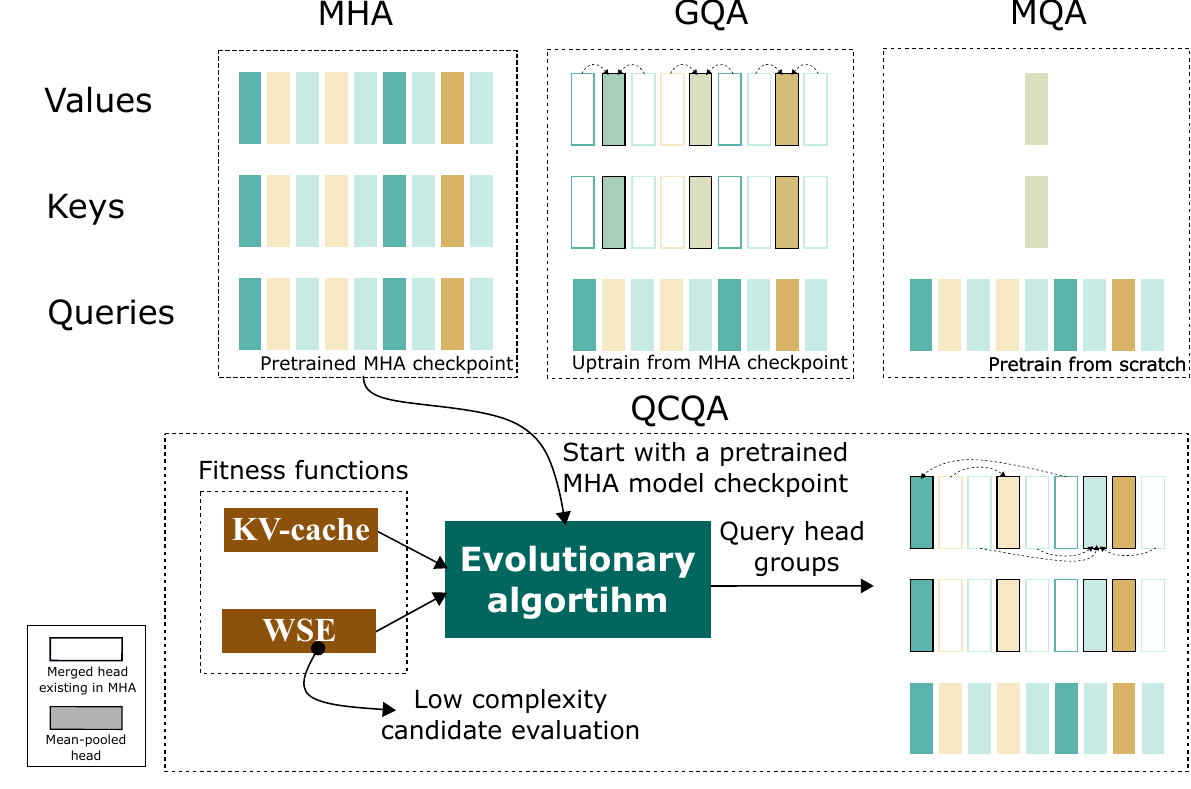}
    
    \caption{Illustration of grouping approaches employed in grouped-query attention (GQA), multi-query attention (MQA), and QCQA. Solid rectangles indicate heads, outlined rectangles indicate a head that was merged, and solid rectangles with a black outline indicate a mean-pooled head. The arrows indicate the merging of heads into one. MQA is trained from scratch with a single key (and value) head. GQA forms groups of subsequent query heads of an MHA checkpoint and mean-pools corresponding key (and value) heads. Using an MHA checkpoint, QCQA can form groups with unequal cardinality by leveraging an evolutionary algorithm and inexpensive fitness functions and then mean-pools corresponding key (and value) heads.  }

    \label{fig:grouped_attentions}
\end{figure}

We propose Quality and Capacity-aware grouped Query Attention (QCQA) to strike an excellent balance between LLM accuracy drop and KV-cache size due to the grouping of query heads. 
As illustrated in Figure \ref{fig:grouped_attentions}, QCQA allows for a fully flexible grouping of query heads in terms of the size of the group and selecting heads in the groups. Beyond forming arbitrary-sized groups, QCQA can constrain input representation to form equal-sized groups.  
Moreover, to eliminate the need for expensive GPU-based LLM accuracy (or perplexity) we devised a simple and computationally inexpensive fitness function, weight-sharing error (WSE), for candidate evaluations in the NSGA-II algorithm. Next, we discuss these formulations in detail below and validate their efficacy in the section \ref{sec:results}.

\subsection{Weight-sharing error (WSE)}
\label{sec:proxy}
Grouped attention techniques introduce errors in attention layers as multiple query heads interact with a single merged key and value head. We developed WSE to indicate the LLM accuracy drop due to the error introduced by the grouping of query heads. 
For a given layer in LLM with $H$ heads, MHA attention computation $A_i$ for $i^{th}$ head with query ($Q_{i}$), key ($K_{i}$), and value ($V_{i}$) features of $d$ dimension is given by,
\begin{equation}
    \label{eq:MHA}
    A_{i} = softmax(\frac{Q_{i} K_{i}^T}{\sqrt{d}}) V_{i}
\end{equation}

Let $K_{G_j}$ and $V_{G_j}$ denote a merged key and value feature heads, respectively, for the $j^{th}$ group $G_j$, then the grouped attention for $i^{th}$ query head $\hat{A_i}$ that belongs to $G_j$ group is given by,

\begin{equation}
    \label{eq:GQA}
    \hat{A_{i}} = softmax(\frac{Q_{i} K_{G_j}^T}{\sqrt{d}}) V_{G_j}
\end{equation}

From equations~(\ref{eq:MHA}) and~(\ref{eq:GQA}) it is clear that the difference between the distributions of $A_i$ and $\hat{A_i}$ leads to accuracy loss in LLMs and by minimizing it accuracy drop for grouped attention techniques can be improved significantly.
Instead of estimating the distance between two attention distributions, \bt{the accuracy drop can be computed more economically} by using key and value head distributions, as given by, 
\begin{equation}
    \hat{L} = \sum_{j=1:P} \sum_{i \in G_j} \big [E[ (K_i - K_{G_j} )^2 ] + E[ (V_i - V_{G_j} )^2 ] \big ] 
\end{equation} 
Where $P$ is the number of groups. 
Since the only source of error is the merging of key and value weight matrices, a simple and inexpensive alternative formulation is,

\begin{equation}
    \label{eq:quality_proxy}
    L = \sum_{j=1:P} \sum_{i \in G_j} \big [E[ (W_{K_i} - W_{K_{G_j}} )^2 ] + E[ (W_{V_i} - W_{V_{G_j}} )^2 ] \big ] 
\end{equation} 

Where $W_{K_{i}}$ and $W_{V_{i}}$ are weight matrices of $i^{th}$ key and value head, respectively. Similarly, $W_{K_{Gj}}$ and $W_{V_{Gj}}$ are mean-pooled key and value head weights belonging to $G_{j}^{th}$ group, respectively. The formulation in equation (\ref{eq:quality_proxy}) is simple and does not depend on input data or actual numerical values of key and value features. 
This formulation bears a precise resemblance to the sum of squared errors (SSE) used in the K-means algorithm.
\bt{Since the mean-pooled head $W_{K_{G_j}}$ or $W_{V_{G_j}}$ are shared with corresponding query heads in the group, we refer to the formulation in equation \ref{eq:quality_proxy} as the \textbf{weight-sharing error (WSE)}.}
Our experiments (section \ref{sec:results}) show a strong relationship between accuracy drop and WSE. Hence, WSE could be reliably used as an alternative to expensive LLM accuracy estimation.





\subsection{Problem formulation}
\label{sec:problem_formulation}
The correct mathematical representation of a candidate is very critical for employing EA for search. A sampling method for the initial random population for EA is dependent on the input representation. Next, we formulate the candidate representation for enabling arbitrary and equal cardinality groups.

\textbf{QCQA with arbitrary cardinality groups (QCQA-AC)} This approach allows EA to form arbitrary-sized groups of query heads by arranging as many heads as possible in a group while keeping accuracy impact minimal.
As a result, this variant offers the most flexibility in forming groups. 
For $H$ number of heads and at most $P$ groups, we adopt the following candidate representation: 
\begin{center}
\centering
$X \in \{\mathbf{0, 1, \ldots, P-1}\}^{H}$ 
\end{center}
In this representation, each $i^{th}$ element of $X$ is the $i^{th}$ head belonging to the group $X[i]$ as illustrated in supplementary Figure \ref{fig:qcqa-ac}.
All such possible values of $X$ will yield a valid and robust input representation for which computation of KV-cache memory requirement and proxy metric is tractable. 
The initial pool of candidates $X$ will start with randomly formed groups where each group index is uniformly sampled. 

\textbf{QCQA with equal cardinality groups (QCQA-EC)} This type of grouping enforces a constraint on the exact number of heads in a group as shown in supplementary Figure \ref{fig:qcqa-ec}. 
The equal cardinality constraint is enforced as follows: 
\begin{center}
\centering
$X \in \{\mathbf{0, 1, \ldots, H-1}\}^{H}$    \\
\end{center}
such that $X[i] \neq X[j]$ for all $i,j \in \{0, 1, \ldots, H - 1\} $ and $ i \neq j$. Each consecutive $m^{th}$ set of $C$ indices of $X$ belong to $m^{th}$ group $G_{m}$ as $\{m\times C : (m+1) \times C \} \in G_{m}$ \quad $\forall m \in \{0, 1, \ldots, P - 1\} $.
Unlike QCQA-AC, QCQA-EC must have all unique elements in its representation $X$ because each entry indicates an index of a head out of H heads. 
As a result, randomly sampling the population of candidates $X$, crossover, and mutation operations will not always result in a valid representation because it can lead to duplicate elements in $X$.
To address this we propose to initialize as $X~=~[0, 1, ..., H-1]$ and randomly sample a population of a pair of two head indices to be swapped in $X$.
For example, randomly select $i^{th}$ and $j^{th}$ groups to perform a swap operation. Let $SP_1 \sim U([$0, 1, \ldots, P-1$])$ and $SP_2 \sim U([$0, 1, \ldots, P-1$])$ be two uniformly sampled indices. A candidate for the initial random population can be obtained by the swap($X$,~$SP_1$,~$SP_2$) operation.
As a result, the population of swap pairs will produce a valid population of $X$ candidates for EA with different possible group arrangements but all groups will have equal cardinality. 



\subsection{QCQA algorithm}
To achieve an optimal tradeoff between KV-cache size and LLM accuracy QCQA implements a two-stage search framework.
As given in algorithm \ref{algo:qcqa_group} the first stage forms groups of query heads for each layer individually. 
The second stage evaluates the impact of applying grouping to a layer on the LLM accuracy. Layers with a high impact on LLM accuracy are retained in their original MHA implementation otherwise the query heads are grouped to minimize KV-cache.
Both stages use WSE and KV-cache computation for calculating the fitness of candidates.


The algorithm \ref{algo:qcqa_group} requires key and value weight matrices and assumes P groups to be formed at most. The groups are formed for each layer individually by leveraging the WSE and KV-cache as fitness functions.  
We found this to result in better tradeoffs than forming groups using a combined representation of all layers. 
Line 3 in algorithm \ref{algo:qcqa_group} implements QCQA-AC, however, for QCQA-EC we replace the line with the corresponding formulation as discussed above.
$ComputeWSE$ function computes the WSE (see equation \ref{eq:quality_proxy}) for a layer according to the grouping information in each sample of the population.
Similarly, the $ComputeKVcache$ function estimates the KV-cache size of a given grouping.
In a generation, the NSGA-II algorithm operates on a given population and computed fitness functions to return next-generation candidates.
Each layer forms different suitable candidates which must be collated together to get wider and granular variation in KV-cache and WSE. For this, we collate two candidates if they belong to the same out of five different percentiles.

The optimal groups of all layers are then fed to algorithm \ref{algo:qcqa} for selecting layers to preserve the grouping information. 
Other layers are retained in the MHA implementation by not forming groups. 
After obtaining optimal groups from algorithm \ref{algo:qcqa_group} (line 1) an initial population is sampled by sampling an array with a binary random variable with equally likely states (line 2). 
The $i^{th}$ entry of $1$ in a candidate of the population indicates that the grouping is preserved for the $i^{th}$ layer otherwise the layer is retained in the MHA implementation. $AccumulateWSE$ function computes and accumulates WSE across all the selected layers given by a candidate in the population.



\begin{algorithm}
    \caption{$QCQAGroups$: Quality and capacity-aware grouping of Query heads in MHA} 
    \begin{algorithmic}[1]
    \Require Number of heads $H$, key and value weight matrices ($W_{K}$ and $W_{V}$), max number of groups $P$, population size $pop\_size$, and number of generations $ngen$.
    \State $groups \leftarrow [[]]\times 5$ \Comment{5 empty lists for candidate collation}
    \For {$layer \leftarrow 1,2,\ldots$}
    \State $pop \leftarrow UniformInteger(0, P-1).sample([pop\_size, H]) $
        \For {$gen \leftarrow 1,2,\ldots,ngen$}
            \State $ fitness.WSE \leftarrow ComputeWSE(pop, W_{K}^{layer}, W_{V}^{layer})$
            \State $fitness.KVcache \leftarrow ComputeKVcache(pop)$                
		\State $pop \leftarrow NSGA-II(pop, fitness)$
        \EndFor
        \For {$percentile \leftarrow enumerate(0,25,50,75,100)$}
		\State $groups[percentile].collate(pop)$
        \EndFor
    \EndFor
    \State \textbf{Return :} $ groups$
    \end{algorithmic} 
    \label{algo:qcqa_group}
\end{algorithm}

\begin{algorithm}
    \caption{QCQA: Selection of layers with multi-head or grouped-head attention} 
    \begin{algorithmic}[1]
    \Require Number of heads $H$, key and value weight matrices ($W_{K}$ and $W_{V}$), max number of groups $P$, population size $pop\_size1$ and $pop\_size2$, number of generations $ngen1$ and $ngen2$, and number of layers $nlayers$.
    \State $groups \leftarrow QCQAGroups(H, W_{K}, W_{V}, P, pop\_size1, ngen1)$
    \State $pop \leftarrow Bernoulli(prob=0.5).sample([pop\_size2, nlayers]) $
    \For {$gen \leftarrow 1,2,\ldots,ngen2$}    
        \State $ fitness.WSE \leftarrow AccumulateWSE(pop, W_{K}, W_{V}, groups)$
        \State $fitness.KVcache \leftarrow ComputeKVcache(pop, groups)$                
        \State $pop \leftarrow NSGA-II(pop, fitness)$
    \EndFor
    \State \textbf{Return :} $ pop, groups$
    \end{algorithmic} 
    \label{algo:qcqa}
\end{algorithm}
\section{Evaluation}
\begin{figure}[!htb]
    \centering
    \includegraphics[width=1.\textwidth]{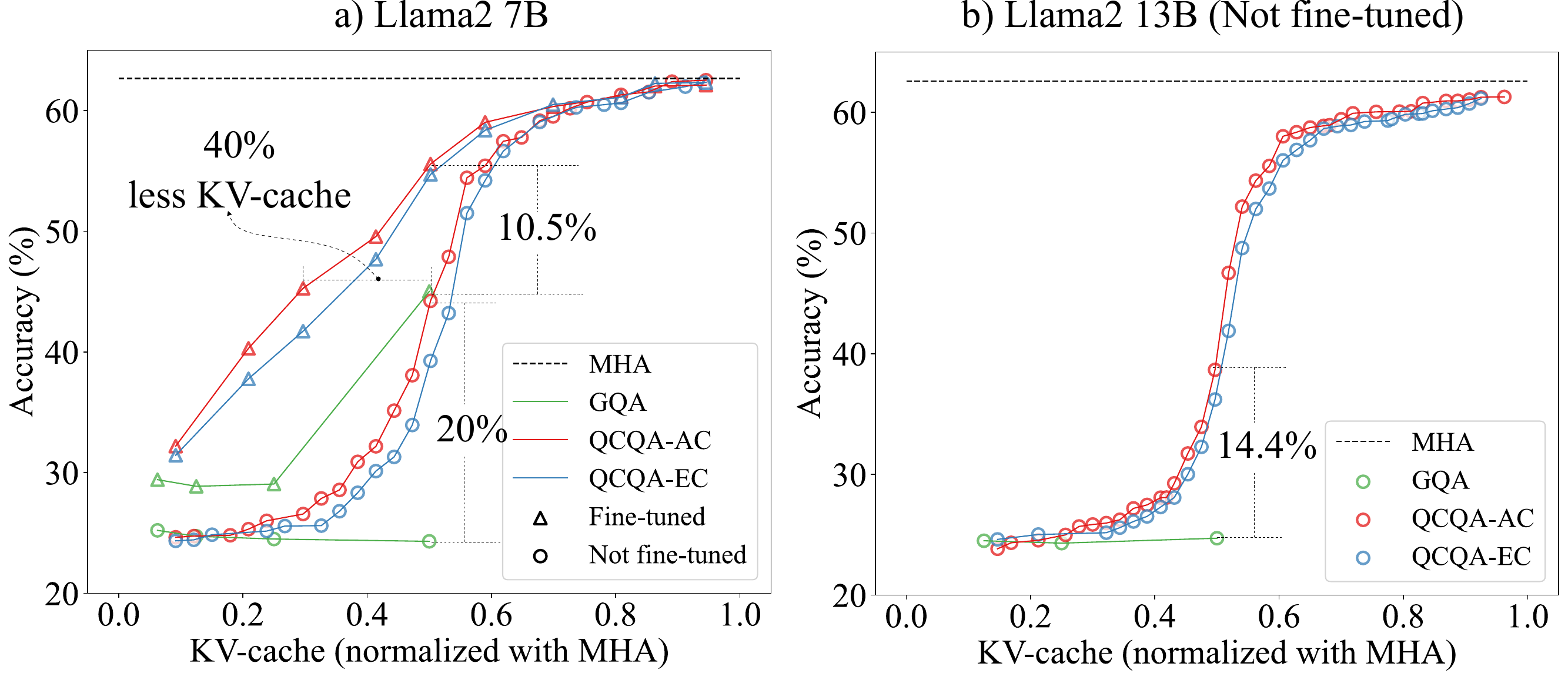}
    \caption{LLM task accuracy at different KV-cache sizes with multiple grouping approaches. Both QCQA-AC and QCQA-EC consistently outperform GQA.}
    \label{fig:acc_kv_size_tradeoff}
\end{figure}
\label{sec:results}
\bt{
We evaluate QCQA on multiple LLM models, tasks, and datasets. From a given pretrained MHA model we derive QCQA and GQA models with different KV-cache sizes. The reported KV-cache sizes are normalized with that of the MHA model. For each of these models, we evaluate LLM task accuracy (average of \textit{Hellaswag} \cite{hellaswag}, \textit{ARC}-\textit{easy} and \textit{challenge} \cite{arc} benchmarks) at different KV-cache sizes before and after fine-tuning. We used the alpaca-cleaned dataset \cite{alapaca-cleaned, alpaca} for fine-tuning.  Further details on our models, datasets, experiments, hyperparameters, and evaluation setup are provided in the supplementary section \ref{sec:exp_details}.}
\subsection{LLM evaluation accuracy analysis}
\bt{
Figure \ref{fig:acc_kv_size_tradeoff} shows LLM task accuracy achieved by QCQA and GQA at different KV-cache sizes. GQA allows for only a few discrete sizes of query groups.
QCQA offers finer granularity of KV-cache sizes due to the flexibility in choosing an arbitrary number of heads per group as well as the number of layers for grouping. Such flexibility can enable the LLM model optimization for any memory budget.}

\textbf{Results without fine-tuning} As shown in Figures \ref{fig:acc_kv_size_tradeoff} (a) and (b), for the KV-cache size of $0.5$, GQA accuracy drops to $24.3\,$\% for Llama2 $7$B and $24.7\,$\% for Llama2 $13$B and QCQA-AC achieves $44.3\,$\% for Llama2 $7$B and $38.6\,$\% for Llama2 $13$B. For both Llama2 model sizes, GQA loses significant accuracy and does not show any improvement in accuracy with increasing KV-cache size, however, QCQA (both AC and EC) shows an excellent tradeoff between KV-cache size and average accuracy. QCQA provides steep accuracy gains for the KV-cache sizes in the interval of [0.3, 0.7]. In the same interval, QCQA-AC outperforms QCQA-EC for both Llama2 model sizes. This demonstrates the efficacy of arbitrary-sized grouping of query heads over that of equal-sized.

\textbf{Results with fine-tuning} Fine-tuned QCQA (both AC and EC) consistently outperforms GQA for all KV-cache sizes as shown in Figure \ref{fig:acc_kv_size_tradeoff} (a). At $0.5$ KV-cache size, QCQA-AC achieves $55.56\,$\% which is $10.55\,$\% higher than GQA ($45.01\,$\%). As evident from the accuracy trend of QCQA-AC and QCQA-EC, arbitrary cardinality provides noticeable accuracy improvements over equal cardinality below $0.7$ KV-cache size. Unlike GQA, QCQA-AC (and EC) both gain in accuracy by a large margin after fine-tuning for lower KV-cache sizes. This allows QCQA-AC to achieve the same accuracy as that of GQA for $40$\% lesser KV-cache size. Compared to MHA, QCQA-AC (and EC) achieve almost $20\,$\% savings in KV-cache with a minor drop ($\approx 2\,$\%) in average accuracy.

Interestingly, at $0.5$ KV-cache size, not fine-tuned QCQA-AC model achieves the same accuracy as that of GQA fine-tuned for 3 epochs. 
\bt{
The benefits of quality and capacity-aware grouping of query heads are clearly visible across different KV-cache sizes. 
QCQA-EC archives $14\,$\% higher accuracy than GQA for the same KV-cache size. 
This shows the effectiveness of our approach even for equal-sized groups.
Accuracy improvement by using query groups of arbitrary cardinality can be significant particularly when fine-tuning is not availed. 
QCQA-AC provides up to $4-8$ \% higher accuracy than QCQA-EC for KV-cache sizes in the mid-range interval [$0.3$, $0.7$].
}
\bt{
\subsection{QCQA evaluation on diverse LLM tasks and benchmarks}
We evaluate QCQA on a range of tasks such as natural language understanding, reasoning, and language generation on different benchmark datasets Hellaswag \cite{hellaswag}, ARC \cite{arc}, MNLI \cite{wang2019glue-mnli}, Winogrande \cite{sakaguchi2019winogrande}, and Wikitext \cite{wiki}. 
Table \ref{tab:more_llm_tasks} shows the accuracy (and perplexity) of QCQA and GQA models at different KV-cache sizes. 
QCQA and GQA models are derived and fine-tuned from the same pretrained checkpoint of the Llama2 $7$B model. 
Across different benchmarks, QCQA achieves similar accuracy (and perplexity) as MHA with a $20$\%  reduction in KV-cache. 
At $0.5$ and $0.25$ KV-cache, the accuracy with QCQA is higher than that of GQA on different benchmarks by ($8\,$\% to $13\,$\%) and ($6\,$\% to $22\,$\%), respectively. QCQA achieves significantly lower perplexity than the GQA model with a similar KV-cache size on the Wikitext dataset. Particularly, the accuracy and perplexity improvements from QCQA are very significant at lower KV-cache sizes compared to GQA.}

\begin{table}[!htb]
    \centering
        \caption{Performance of QCQA grouping compared to GQA on diverse benchmarks. The KV-cache size is normalized by that of the original capacity required for MHA.}
        \begin{tabular}{ ccccccc } 
         \hline
         Model &  \textbf{KV-cache} & \textbf{Hellaswag} & \textbf{ARC} (\%) &  \textbf{MNLI} & \textbf{Winogrande} & \textbf{WikiText} \\ 
           &  \textbf{size} & (\%) & (easy | challenge) & (\%) & (\%) & PPL \\
         \hline
         MHA & 1.00 & 60.97 & 78.57 | 49.57 & 40.98 & 68.90 & 10  \\ 
         QCQA-AC & 0.80 & 60.47 & 76.47 | 46.50 & 44.24 & 68.03 & 11 \\ 
         QCQA-AC & 0.50 & 55.52 & 67.42 | 39.07 & 43.38 & 65.11 & 22  \\ 
         QCQA-AC & 0.20 & 39.97 & 54.63 | 26.28 & 34.60 & 50.98 & 57  \\ 
         GQA & 0.50 & 45.33 & 58.31 | 31.48 & 36.99 & 52.80 & 98 \\
         GQA & 0.25 & 32.67 & 33.59 | 20.90 & 36.00 & 51.85 & 181 \\
         \hline
        \end{tabular}    
    \label{tab:more_llm_tasks}
\end{table}

\begin{figure}[!htb]
    \centering
    \includegraphics[width=.9\textwidth]{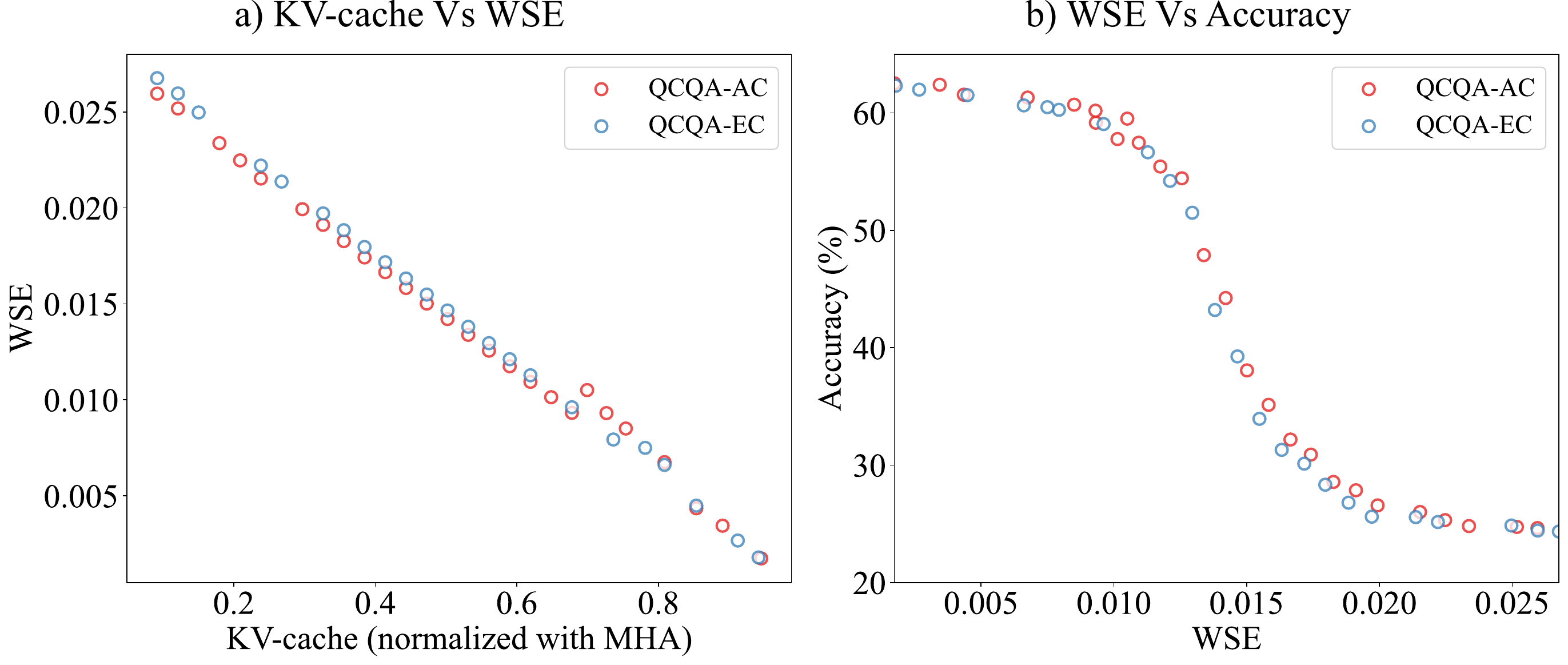}
    \caption{Relation between WSE, KV-cache, and LLM evaluation accuracy. }
    \label{fig:acc_qloss_corr}
\end{figure}
\subsection{WSE, KV-cache, and LLM evaluation accuracy}
\label{sec:result-wse-kv-acc}
\bt{As discussed in section \ref{sec:methodology}, QCQA uses an evolutionary algorithm to obtain quality and capacity-aware grouping of query heads. To evaluate different candidates in the NSGA-II algorithm we proposed WSE as an inexpensive fitness function instead of LLM accuracy. Figure \ref{fig:acc_qloss_corr} shows the relation between WSE, KV-cache size, and LLM accuracy obtained from the Llama2 7B model for QCQA-AC and QCQA-EC.}

\bt{Figure \ref{fig:acc_qloss_corr} (a) shows that the WSE is inversely proportional to the KV-cache size. This is because the KV-cache size reduces as more heads are grouped for which WSE increases. Higher WSE indicates a larger distance between the distributions of mean-pooled weights and the corresponding weights of the KV heads in the group. Consequently, higher WSE indicates lower accuracy. Though the relation between accuracy and WSE is non-linear, there is a monotonic decrease in the accuracy with increasing WSE as shown in Figure \ref{fig:acc_qloss_corr} (b) which conforms with our assumptions and formulation in the section \ref{sec:proxy}. This validates that WSE can serve as a reliable inexpensive alternative for accuracy drop.} We have explored the efficacy of QCQA-AC for Llama2 13B and OPT models by studying the Pareto charts given in supplementary Figures \ref{fig:pareto-llama2} and \ref{fig:pareto-opt}. The number of groups $P$ in QCQA-AC indicates that the NSGA-II algorithm will not form more than $P$. The WSE for the QCQA approach is consistently better than GQA for different numbers of groups, demonstrating the scalability of our approach to other LLMs.


\section{Related works}
Previous research has explored quantization for relaxing KV-cache memory requirements along with LLM weights \cite{hooper2024kvquant, frantar2023optq}. 
Beyond quantization, several eviction strategies were explored to remove KV features corresponding to non-important tokens \cite{liu2023scissorhands, zhang2023h2o, anagnostidis2023dynamic}. 
In \cite{delcorro2023skipdecode} researchers explore token-level early exit strategy for each token in a batch and sequence length to skip the execution of some layers.
The two most closely related works are \cite{ainslie2023gqa, shazeer2019fast_mqa} which proposed shrinking the head dimension of key and value features by merging multiple heads into one. 
In \cite{javadi2023gqkva} authors proposed the grouping of query, key, and value heads independently to reduce the attention computation for vision transformer networks. 
Dynamic memory compression (DMC) proposed in \cite{nawrot2024dynamic} offers online compression of KV-cache memory at inference time by accumulating some of the key and value features corresponding to new tokens. DMC applies different compression rates for different heads and layers. 
Another similar approach to ours is Autonac \cite{autonac}, a proprietary NAS tool by Deci.ai that proposes the use of NAS to arrange heads with different numbers of groups across layers but maintaining equal-sized groups in each layer. For example, some layers may have GQA with 2 groups and others GQA with 8 groups. However, the objective function used in their NAS formulation is LLM evaluation accuracy (page 16 in white paper \cite{autonac}) which often can be expensive to evaluate. We expect Autonac results to be closest to QCQA-EC but as shown in section \ref{sec:results}, QCQA-AC outperforms QCQA-EC in most cases.


In weight sharing NAS work \cite{cummings2022hardwareaware, kundu2024cimnet} propose the use of predictor functions to estimate the fitness of a candidate during evaluations of EA. 
A trainable predictor-based fitness function may be useful as an alternative to WSE, albeit at the cost of expensive GPU runs for initial training.

\section{Discussion and Limitations  }
\label{sec:discuss}

Allowing flexibility in quality-aware grouping of query heads provides a significantly better tradeoff between KV-cache size and LLM evaluation accuracy.
The WSE formulation and the evolutionary search paradigm used in this work for quality-aware grouping of heads can be relevant to other techniques for KV-cache optimization such as using low precision for attention heads. 
Since QCQA groups such that weight distribution has minimal distance before and after pooling, we expect that QCQA would need lesser uptraining efforts than GQA to obtain accuracy comparable to MHA.
Though we spent major efforts in studying QCQA, we leave some scope for future exploration such as 1) weighted summation of terms in equation (\ref{eq:quality_proxy}), 2) in algorithm \ref{algo:qcqa} allowing groups of either key or value or both in a layer, and 3) other distance metrics instead of SSE in WSE.

This body of work currently only applies to autoregressive inference of LLMs. 
We have demonstrated fine-tuned results on the Llama2 $7$B model but due to prohibitive resource costs, we were unable to evaluate on larger models.
To reduce the overall computational complexity of the evolutionary search, we did not attempt nested NSGA-II implementation for combining algorithms \ref{algo:qcqa_group} and \ref{algo:qcqa}.
Due to reliance on pretrained model checkpoints, QCQA cannot applied when LLMs must be trained from scratch. 

 

\section{Broader Impacts} 
We anticipate that our work will not result in any negative societal impacts. 
Our work leverages an evolutionary algorithm that can be run on CPU-based hardware alone. 
QCQA derives LLM models optimized for KV-cache with minimal impact on the accuracy.
With a few hours of fine-tuning efforts QCQA further gains in accuracy by a large margin. 
This eliminates the need for expensive pretraining or uptraining of LLM models, saving significantly in $CO_2$ emissions.
The WSE formulation opens new directions for research in weight-sharing NAS to eliminate the need for expensive candidate evaluations.
Memory and power-constrained devices can significantly benefit from lower KV-cache overhead and higher LLM accuracy.
Overall we think that QCQA will have a broadly positive impact by making LLMs with high accuracy and optimized KV-cache widely and easily accessible.

\section{Conclusion}
We have proposed quality and capacity-aware grouped query attention (QCQA) that achieves an optimal tradeoff between KV-cache memory requirements and LLM accuracy for autoregressive inference of LLMs. 
Unlike GQA, QCQA uses an evolutionary algorithm \bt{(NSGA-II)} to form arbitrary-sized groups of query heads of a pretrained LLM checkpoint. 
We demonstrated the efficacy of forming arbitrary and equal cardinality groups of query heads over GQA. 
We devised a simple and computationally inexpensive fitness function, WSE, for using the NSGA-II algorithm for multi-objective optimization to get an optimal tradeoff. 
Using \bt{WSE as an alternative to LLM accuracy we enable} the estimation of candidate fitness on CPU, thus avoiding the expensive GPU-based LLM evaluations.
In the absence of fine-tuning QCQA achieves $\mathbf{20}$\% more accuracy than GQA for the same KV-cache size. \bt{After} fine-tuning \bt{GQA and QCQA both, QCQA continues to outperform GQA with $\mathbf{10.55}$\% higher accuracy at similar KV-cache size. 
Furthermore, QCQA requires $\mathbf{40}\,$\% lesser KV-cache than GQA to attain the same accuracy. 
We have presented a new paradigm for KV-cache optimization by using WSE and evolution search for quality and capacity-aware grouping of query heads and layers. 
Our approach in part or whole can be relevant to other techniques for KV-cache optimization.
}



\bibliography{references}
\bibliographystyle{unsrt} 






\newpage
\appendix
\section{Appendix / supplemental material}

\subsection{Illustration of QCQA grouping using arbitrary and equal cardinality}
\begin{figure}[!htb]
    \centering
    \includegraphics[width=1.\textwidth]{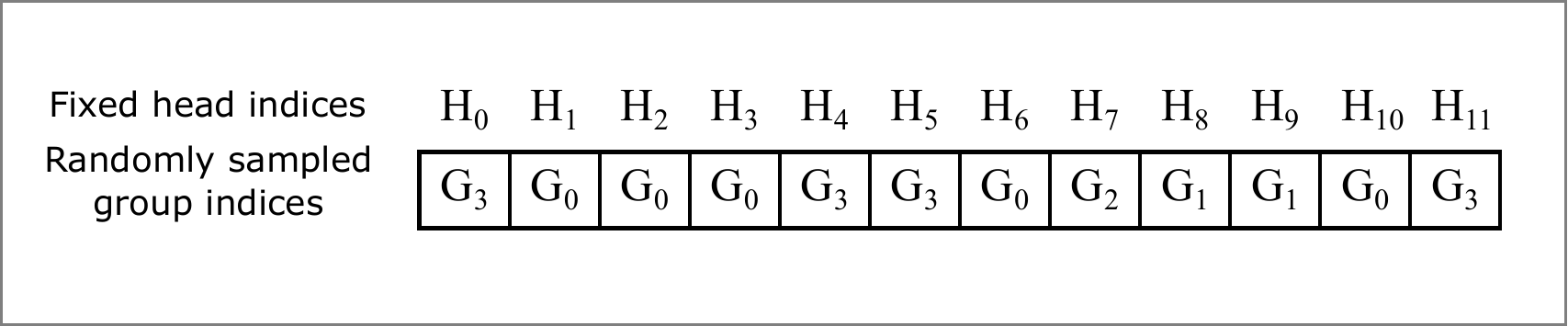}
    \caption{Illustration of QCQA grouping approach for enabling arbitrary cardinality. Each candidate $X$ is a vector of $H$ elements and each element indicates a head index. The value at each head index indicates the group to which the head belongs.  }
    \label{fig:qcqa-ac}
\end{figure}

As discussed in section \ref{sec:methodology}, QCQA-AC forms arbitrary-sized groups of query heads illustrated in Figure \ref{fig:qcqa-ac}. Each candidate is a vector of size $H$ and an index in the vector indicates the index of the head. The value at each index of the vector indicates the group to which the index belongs.

\begin{figure}[!htb]
    \centering
    \includegraphics[width=1.\textwidth]{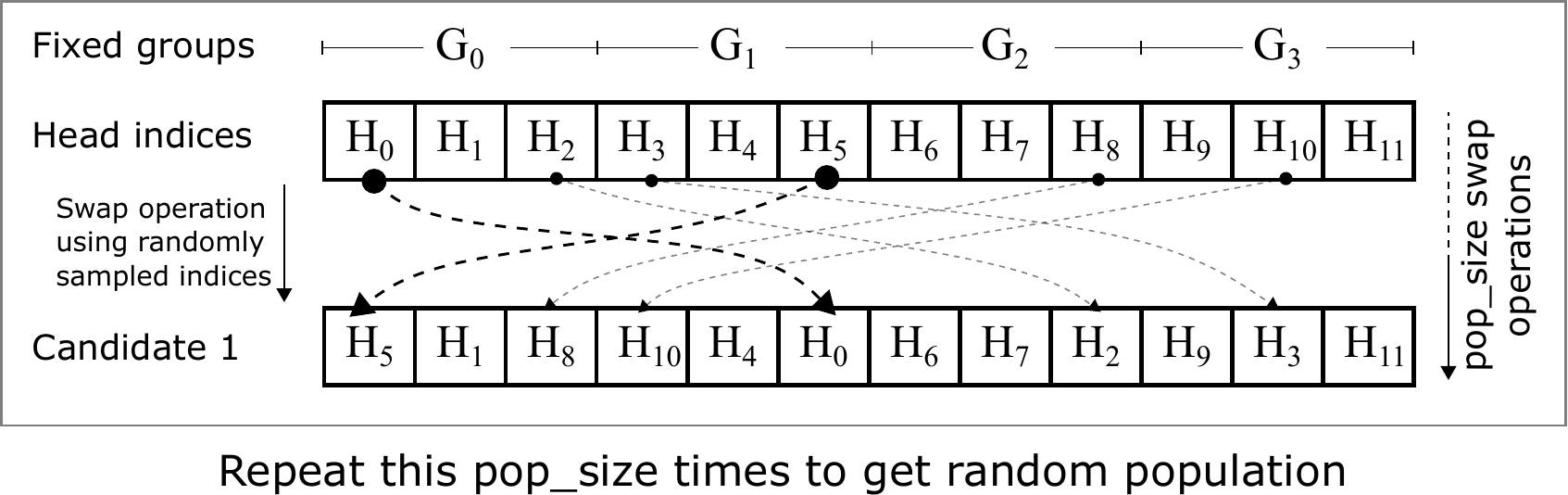}
    \caption{Illustration of forming groups of equal cardinality in QCQA. Each candidate $X$ is a vector of $H$ elements. Unlike QCQA-AC, the groups are fixed as consecutive $P$ sets of the elements of $X$. The value at each element of $X$ indicates the head index. Multiple swapping operations are applied at once to obtain a random sample of $X$. Repeating this for the number of population size ($pop\_size$) times provides the initial random population for the evolutionary algorithm.  }
    \label{fig:qcqa-ec}
\end{figure}

QCQA-EC enforces the equal-sized group criteria in the candidate representation and this is illustrated in Figure \ref{fig:qcqa-ec}. In QCQA-EC the candidate is a vector of length $X$ but, unlike QCQA-AC, a set of indices in the vector represents a group. To ensure equal-size groups, subsequent $C$ sets of indices are assigned a group. The value at an index in the vector gives the head that belongs to a group. 

\subsection{Evaluation setup}
\label{sec:exp_details}
We use \textit{PyTorch}-based \cite{paszke2019pytorch} LLM fine-tuning and evaluation framework \textit{torchtune} \cite{torchtune} for our evaluations. It supports a wide range range of tasks and datasets. The \textit{torchtune} natively supports GQA. We have extended \textit{torchtune} to support QCQA for LLM inference and fine-tuning. 

\textbf{Models and Datasets} We use Llama2 $7$B and Llama2 $13$B multi-headed attention (MHA) models for our evaluations. We fine-tuned Llama2 $7$B on the alpaca-cleaned dataset \cite{alapaca-cleaned, alpaca} for 3 epochs with a learning rate of $2 \times 10^{-5}$ and a mini-batch size of 2. We use a pretrained MHA model checkpoint to derive QCQa and GQA models. All fine-tuning is performed only on the alpaca-cleaned dataset.

For NSGA-II experiments, we use \textit{PyMOO} \cite{pymoo} python module for implementing multi-objective optimization. 
The NSGA-II algorithm provides an optimal grouping of heads and this information is further fed to \textit{PyTorch}-based \cite{paszke2019pytorch} LLM fine-tuning framework \textit{torchtune} \cite{torchtune}. 

We perform hyperparameter tuning of the NSGA-II algorithm for crossover and mutation probabilities, initial population size, and the number of generations to evaluate before reaching termination criteria. The Llama2 models are representative of the recent state-of-the-art LLMs in research and we choose to conduct most of our experiments on $7\,$B and $13\,$B versions of this model. 
We have explored the scalability of our approach to the OPT ($350\,$M and $6.7\,$B) by applying the QCQA algorithm to obtain optimal grouping for Query heads. 
All other hyperparameters were set to their default settings both in \textit{PyMoo} and \textit{torchtune}.
We did not perform fine-tuning on Llama2 $13$B and OPT models but we report the accuracy or WSE for these models.
In our results, we report normalized KV-cache for all models. Normalization is performed by dividing a KV-cache size value by that of the KV-cache size of the MHA model $\frac{2\times B\times T\times P\times d}{2\times B\times T\times H\times d} = \frac{P}{H}$. The average accuracy in Figures \ref{fig:acc_kv_size_tradeoff} and \ref{fig:acc_qloss_corr} is the average accuracy of \textit{Hellaswag} \cite{hellaswag} and both the \textit{easy} and \textit{challenge} sets of \textit{ARC} \cite{arc} benchmarks. We report individual accuracy on more tasks and benchmarks in table \ref{tab:more_llm_tasks}.

We have used NVIDIA RTX 3090 and A6000 GPUs in all our fine-tuning simulations. NSGA-II-based simulations were run on Intel CPUs. Each fine-tuning run takes approximately 5-8 hours per epoch depending on the GPU. We have used the \textit{lm-evaluation-harness} \cite{eval-harness} framework for LLM evaluation accuracy experiments. Each evaluation takes 0.5-3 hours depending on the GPU.

\subsection{Pareto charts on KV-cache and WSE tradeoff}
\label{sec:more_pareto}
\begin{figure}[!htb]
    \centering
    \includegraphics[width=0.9\textwidth]{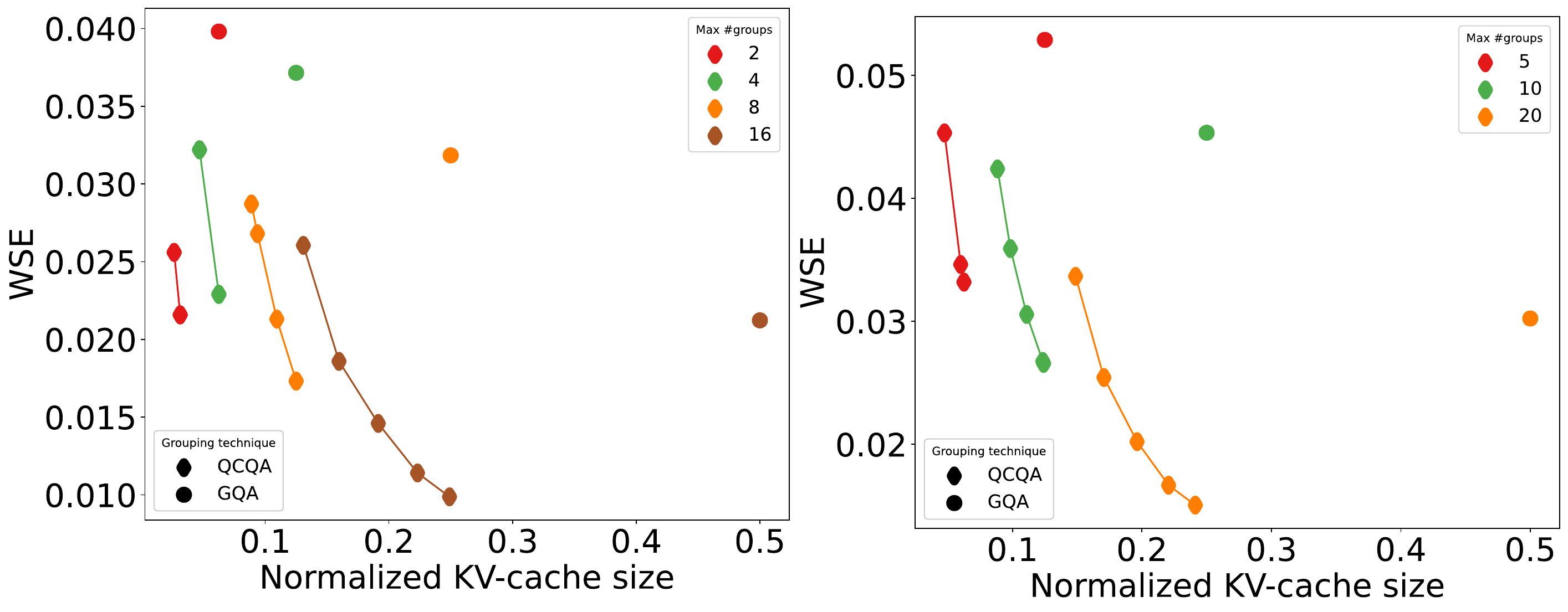}
    \caption{Pareto charts for Llama2 models with size 7B (left) and 13B (right) obtained using QCQA-AC. Legends in color indicate the maximum number of groups to be formed. For GQA this is equal to the number of groups. For QCQA, the NSGA-II algorithm will groups equal to or lower than this number. Each point is the mean of 10 independent runs. The error bars plotted are very small (at least 3 orders of magnitude smaller than the mean) so not visible on the graph. }
    \label{fig:pareto-llama2}
\end{figure}

\begin{figure}[!htb]
    \centering
    \includegraphics[width=0.9\textwidth]{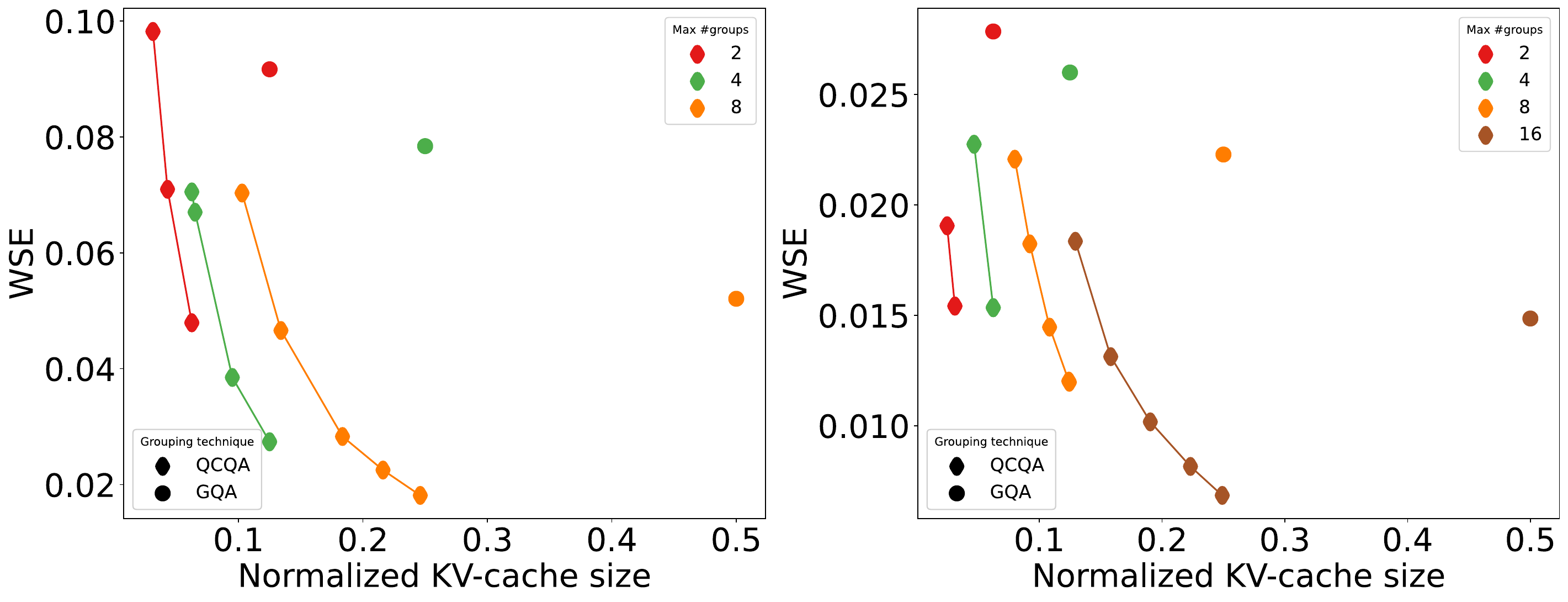}
    \caption{Pareto charts for OPT models with size 350M (left) and 6.7B (right) obtained using QCQA-AC. Legends in color indicate the maximum number of groups to be formed. For GQA this is equal to the number of groups. For QCQA, the NSGA-II algorithm will groups equal to or lower than this number. Each point is the mean of 10 independent runs. The error bars plotted are very small (at least 4 orders of magnitude smaller than the mean) so not visible on the graph.}
    \label{fig:pareto-opt}
\end{figure}

Supplementary Figures \ref{fig:pareto-llama2} and \ref{fig:pareto-opt} show the output of the NSGA-II algorithm using arbitrary grouping to optimize for KV-cache and proxy metric.
The legends in color indicate the at most number of groups explored by the NSGA-II algorithm. 
For example, for at most 8 groups, the NSGA-II algorithm could yield multiple candidates representing different numbers of groups (\textbf{not exceeding 8}) with distinct cardinality in all layers.
For all the models and the number of groups, QCQA substantially outperforms the corresponding GQA implementation. 
The QCQA algorithm yields lower proxy metric values for the same KV-cache sizes. 
In some cases, QCQA candidates with lower KV-cache requirements (due to a lower number of groups per layer) can achieve lower proxy metric values.
\label{sec:more_llms}

\end{document}